\documentclass{article}

\usepackage[numbers,sort&compress]{natbib}
\usepackage[utf8]{inputenc} %
\usepackage[T1]{fontenc}    %
\usepackage{hyperref}       %
\usepackage{url}            %
\usepackage{booktabs}       %
\usepackage{amsfonts}       %
\usepackage{nicefrac}       %
\usepackage{microtype}      %
\usepackage{outlines}
\usepackage[utf8]{inputenc}
\usepackage{amsmath, latexsym}
\usepackage{amscd}
\usepackage{amsbsy}
\usepackage{amssymb}
\usepackage{amsthm}
\usepackage{todonotes}
\usepackage{comment} 
\usepackage{nicefrac}
\usepackage{shortcuts}
\newcommand{\pr}{\mathbb{P}}
\usepackage{amsthm,amsmath,amssymb,enumitem}
\theoremstyle{plain}

\newtheorem{proposition}{Proposition}

\theoremstyle{definition}
\newtheorem{assumption}{Assumption}

\usepackage{mathrsfs}
\usepackage{mathtools}

\usepackage{natbib}
\newcommand{\tpr}{\op{TPR}}
\newcommand{\tnr}{\op{TNR}}
\newcommand{\tp}{\op{TPR}}
\newcommand{\tn}{\op{TNR}}

\usepackage{mathtools}
\usepackage[capitalise]{cleveref}
\Crefname{assumption}{Assumption}{Assumptions}

\title{Assessing Disparate Impacts of Personalized Interventions: Identifiability and Bounds}

\author{%
Nathan Kallus\\
Cornell University\\
\texttt{kallus@cornell.edu}
\and
Angela Zhou\\
Cornell University\\
\texttt{az434@cornell.edu}
}
\date{}

\begin{document}

\maketitle

\begin{abstract}
Personalized interventions in social services, education, and healthcare leverage individual-level causal effect predictions in order to give the best treatment to each individual or to prioritize program interventions for the individuals most likely to benefit. While the sensitivity of these domains compels us to evaluate the fairness of such policies, we show that actually auditing their disparate impacts per standard observational metrics, such as true positive rates, is impossible since ground truths are unknown. Whether our data is experimental or observational, an individual's actual outcome under an intervention different than that received can never be known, only predicted based on features. We prove how we can nonetheless point-identify these quantities under the additional assumption of monotone treatment response, which may be reasonable in many applications. We further provide a sensitivity analysis for this assumption by means of sharp partial-identification bounds under violations of monotonicity of varying strengths. We show how to use our results to audit personalized interventions using partially-identified ROC and xROC curves and demonstrate this in a case study of a French job training dataset.
\end{abstract}

\section{Introduction}\label{sec:intro}
The expanding use of predictive algorithms in the public sector for risk assessment 
has sparked recent concern and study of fairness considerations 
\citep{propublica,bs14,barocas-hardt-narayanan}. One critique of the use of predictive risk assessment argues that the discussion should be reframed to instead focus on the role of \textit{positive interventions} in distributing beneficial resources, such as directing pre-trial services to prevent recidivism, rather than in meting out pre-trial detention based on a risk prediction \cite{bdivz17}; or using risk assessment in child welfare services to provide families with additional childcare resources rather than to inform the allocation of harmful suspicion \cite{shroff2017predictive,eubanks2018automating}.
However, due to limited resources, interventions are necessarily targeted. Recent research specifically investigates the use of models that predict an intervention's benefit in order to efficiently target their allocation, such as in developing triage tools to target homeless youth \cite{triage-rice-13,kube-das-19}.
Both ethics and law compel such personalized interventions to be fair and to avoid disparities in how they impact different groups defined by certain protected attributes, such as race, age, or gender.

The delivery of interventions to better target those individuals deemed most likely to respond well, even if a prediction or policy allocation rule does not have access to the protected attribute, might still result in disparate impact (with regards to social welfare) for the same reasons that these disparities occur in machine learning classification models \citep{chen-js18}. (See \cref{apx-substantive} for an expanded discussion on our use of the term ``disparate impact.'') However, in the problem of personalized interventions, the ``fundamental problem of causal inference,'' that outcomes are not observed for interventions not administered, poses a fundamental challenge for evaluating the fairness 
of any intervention allocation rule, as the 
true ``labels'' of intervention efficacy
of any individual are never observed in the dataset. 
Metrics commonly assessed in the study of fairness in machine learning, such as group true positive and false positive rates, are therefore conditional on potential outcomes which are not observed in the data and therefore cannot be computed as in standard classification problems. 

The problem of personalized policy learning has surfaced in econometrics and computer science \cite{manski05,kt15}, gaining renewed attention alongside recent advances in causal inference and machine learning \cite{athey17,wager17,dell2014}. In particular, \cite{bd12} analyze optimal treatment allocations for malaria bednets with nonparametric plug-in estimates of conditional average treatment effects, accounting for budget restrictions; \cite{davis-heller} use the generalized random forests method of \cite{wager2017estimation} to evaluate heterogeneity of causal effects in a program matching at-risk youth in Chicago with summer jobs on outcomes and crime; and \cite{kube-das-19} use BART \citep{hill2011bayesian} to analyze heterogeneity of treatment effect for allocation of homeless youth to different interventions, remarking that studying fairness considerations for algorithmically-guided interventions is necessary.

In this paper, we address the challenges of assessing the disparate impact of such personalized intervention rules in the face of unknown ground truth labels. We show that we can actually obtain point identification of common observational fairness metrics under the assumption of \textit{monotone treatment response}. We motivate this assumption and discuss why it might be natural in settings where interventions only either help or do nothing. Recognizing nonetheless that this assumption is not actually testable, we show how to conduct sensitivity analyses for fairness metrics. In particular, we show how to obtain sharp partial identification bounds on the metrics of interest as we vary the strength of violation of the assumption. We then show to use these tools to visualize disparities using partially identified ROC and xROC curves. We illustrate all of this in a case study of personalized job training based on a dataset from a French field experiment.

\section{Problem Setup}

We suppose we have data on individuals $(X,A,T,Y)$ consisting of:
\begin{itemize}
  \item Prognostic features $X \in \mathcal{X}$, upon which interventions are personalized;
  \item Sensitive attribute $A\in\mathcal A$, against which disparate impact will be measured;
  \item Binary treatment indicator $T\in \{0,1\}$, indicating intervention exposure; and
  \item Binary response outcome $Y\in\{0,1\}$, indicating the benefit to the individual.
\end{itemize}
Our convention is to identify $T=1$ with an active intervention, such as job training or a homeless prevention program, and $T=0$ with lack thereof.
Similarly, we assume that a positive outcome, $Y = 1$, is associated with a beneficial event for the individual, e.g., successful employment or non-recidivation.
Using the Neyman-Rubin potential outcome framework \citep{ir15}, we let $Y(0), Y(1) \in \{0,1\}$ denote the potential outcomes of each treatment. 
We let the observed outcome be the potential outcome of the assigned treatment, $Y=Y(T)$, encapsulating non-interference and consistency assumptions, also known as SUTVA \citep{rubin1980randomization}. Importantly, for any one individual, we never \emph{simultaneously} observe $Y(0)$ and $Y(1)$. This is sometimes termed the fundamental problem of causal inference.
We assume our data either came from a randomized controlled trial (the most common case) or an unconfounded observational study so that the treatment assignment is ignorable, that is, $Y(1), Y(0) \indep T \mid X,A$.

When both treatment and potential outcomes are binary, we can exhaustively enumerate the four possible realizations of potential outcomes
as $(Y(0),Y(1))\in\{0,1\}^2$. We call units with $(Y(0),Y(1))=(0,1)$ responders, $(Y(0),Y(1))=(1,0)$ anti-responders, and $Y(0)=Y(1)$ non-responders.
Such a decomposition is also common in instrumental variable analysis \cite{angrist96} where the binary outcome is take-up of treatment with the analogous nomenclature of compliers, never-takers, always-takers, and defiers. In the context of talking about an actual outcome, following \citep{manski1997monotone}, we replace this nomenclature with the notion of response rather than compliance.
We remind the reader that due to the fundamental problem of causal inference, response type is \emph{unobserved}.

We denote the conditional probabilities of each response type by
$$p_{ij}=p_{ij}(X,A)=\mathbb P(Y(0)=i,Y(1)=j\mid X,A).$$ By exhaustiveness of these types,
$p_{00}+p_{01}+p_{10}+p_{11}=1$. (Note $p_{ij}$ are \emph{random variables}.)

We consider evaluating the fairness of a personalized intervention policy $Z=Z(X,A)\in\{0,1\}$, which assigns interventions based on observable features $X,A$ (potentially just $X$).
Note that by definition, the intervention has zero effect on non-responders, negative effect on anti-responders, and a positive effect only on responders. Therefore, in seeking to benefit individuals with limited resources, the personalized intervention policy should seek to target only the responders. Naturally, response type is unobserved and the policy can only mete out interventions based on observables.

In classification settings, minimum-error classifiers on the efficient frontier of type-I and -II errors are given by Bayes classifiers that threshold the probability of a positive label. In personalized interventions, policies that are on the efficient frontier of social welfare (fraction of positive outcomes, $\Prb{Y(Z)=1}$) and program cost (fraction intervened on, $\Prb{Z=1}$) are given by thresholding ($Z=\indic{\tau\geq\theta}$) the \textit{conditional average treatment effect} (CATE):
\begin{align*}
\tau=\tau(X,A)&=\Efb{Y(1)-Y(0)\mid X,A} =p_{01}-p_{10}\\&= \mathbb P(Y=1 \mid T=1, X, A) - \mathbb P(Y=1 \mid T=0, X, A ),
\end{align*} 
where the latter equality follows by the assumed ignorable treatment assignment. Estimating $\tau$ from unconfounded data using flexible models has been the subject of much recent work \citep{wager2017estimation,shalit-johansson-sontag-17,hill2011bayesian}.

We consider observational fairness metrics in analogy to the classification setting, where the ``true label'' of an individual is their \textit{responder status}, $R = \indic{Y(1)>Y(0)}$. We define the analogous true positive rate and true negative rate for the intervention assignment $Z$, conditional on the (unobserved) events of an individual being a responder or non-responder, respectively:
\begin{equation}\label{eq:tprdef}\begin{aligned}
\tpr_a&=\mathbb P(Z=1\mid A=a,Y(1)>Y(0)),\\
\tnr_a&=\mathbb P(Z=0\mid A=a,Y(1)\leq Y(0)).
\end{aligned}\end{equation}

\subsection{Interpreting Disparities for Personalized Interventions} 
The use of predictive models 
to deliver interventions 
can induce disparate impact if responding (respectively, non-responding) individuals of different groups receive the intervention at disproportionate rates under the treatment policy. 
This can occur even with efficient policies that threshold the true CATE $\tau$ and can arise from the disparate predictiveness of $X,A$ of response type (i.e., how far $p_{ij}$ are from $0$ and $1$). This is problematic because the choice of features $X$ is usually made by the intervening agent (e.g., government agency, etc.).

We discuss one possible interpretation of TPR or TNR disparities in this setting when the intervention is the bestowal of a benefit, like access to job training or case management.
From the point of view of the intervening agent, there are specific program goals, such as employment of the target individual within 6 months.
Therefore, false positives are costly due to program cost and false negatives are missed opportunities.
But outcomes also affect the individual's utility.
Discrepancies in TPR across values of $A$ are of concern since they suggest that the needs of those who could actually benefit from intervention (responders) in one group are not being met at the same rates as in other groups. Arguably, for benefit-bestowing interventions, TPR discrepancies are of greater concern.
Nonetheless, from the point of view of the individual, the intervention may always grant some positive resource (e.g., from the point of view of well-being), regardless of responder status, since it corresponds to access to a good (and the individual can gain other benefits from job training that may not necessarily align with the intervener's program goals, such as employment in 1 year or personal enrichment). If so, then TNR discrepancies across values of $A$ imply a ``disparate benefit of the doubt'' such that the policy disparately over-benefits one group over another using the limited public resource without the cover of advancing the public program's goal, which may raise fairness and envy concerns, especially since this ``waste'' is at the cost of more slots for responders. 

Beyond assessing disparities in TPR and TNR for one fixed policy, we will also use our ability to assess these over varying CATE thresholds in order to compute xAUC metrics \citep{kz-19} in \cref{sec-group-disparities}. These give the disparity between the probabilities that a non-responder from group $a$ is ranked above a responder from group $b$ and vice-versa. Thus, they measure the disproportionate access one group gets relative to another in \emph{any} allocation of resources that is non-decreasing in CATE.

We emphasize that the identification arguments and bounds that we present on fairness metrics are primarily intended to facilitate the \textit{assessment} of disparities, which may require further inquiry as to their morality and legality, not necessarily to promote statistical parity via adjustments such as group-specific thresholds, though that is also possible using our tools. We defer a more detailed discussion to \cref{sec-discussion} and re-emphasize that assessing the distribution of outcome-conditional model errors are of central importance both in machine learning \cite{hardt2016equality,barocas-hardt-narayanan,mitchell-potash-barocas} and in the economic efficiency of targeting resources \cite{pmt-2016,pmt-2016-brown,berger}.

\section{Related Work}

\cite{madras-creager-pz19} consider estimating joint treatment effects of race and treatment under a deep latent variable model to reconstruct unobserved confounding. For evaluating fairness of policies derived from estimated effects, they consider the gap in population accuracy $\mathrm{Acc}_a = \Prb{Z = Z^*\mid A=a}$, where $Z^*= \mathbb{I}[\tau(X) > 0]$ is the (identifiable) optimal policy. In contrast, we highlight the unfairness of even optimal policies and focus on outcome-conditional error rates (TPR, TNR), where the non-identifiability of responder status introduces challenges regarding identifiability.

The issue of model evaluation under the censoring problem of selective labels has been discussed in situations such as pretrial detention, where detention censors outcomes \cite{lkllm17,kz18-2}. Sensitivity analysis is used in \cite{jsfg18} to account for possible unmeasured confounders. The distinction is that we focus on the targeted delivery of interventions with unknown (but estimated) causal effects, rather than considering classifications that induce one-sided censoring but have definitionally known effects.

Our emphasis is distinct from other work discussing fairness and causality that uses graphical causal models to decompose predictive models along causal pathways and assessing the normative validity of path-specific effects \cite{klrs17,hardtscholkopf-disc17}, such as the effect of {probabilistic} hypothetical interventions on race variables or other potentially immutable protected attributes. When discussing treatments, we here consider interventions corresponding to allocation of concrete resources (e.g., give job training), which are in fact physically manipulable by an intervening agent.
The correlation of the intervention's \emph{conditional average} treatment effects by, say, race and its implications for downstream resource allocation are our primary concern. 

There is extensive literature on partial identification in econometrics, e.g. \cite{manski2003partial}. In contrast to previous work that analyzes partial identification of average treatment effects when data is confounded and using monotonicity to improve precision \citep{manski2003partial,balke1997bounds,beresteanu2012partial}, we focus on unconfounded (e.g., RCT) data and achieve full identification by assuming monotonicity and consider sensitivity analysis bounds for \textit{nonlinear} functionals of partially identified sets, namely, true positive and false positive rates. 

\section{Identifiability of Disparate Impact Metrics}\label{sec-fairness-id}

Since the definitions of the disparate impact metrics in \cref{eq:tprdef} are conditioned on an unobserved event, such as the response event $Y(1) > Y(0)$, they actually cannot be identified from the data, even under ignorable treatment. That is, the values of $\tpr_a,\tnr_a$ can vary even when the joint distribution of $(X,A,T,Y)$ remains the same, meaning the data we see cannot possibly tell us about the specific value of $\tpr_a,\tnr_a$.
\begin{proposition}\label{lemma:unidentifiable}
$\tpr_a,\tnr_a$ (or discrepancies therein over groups) are generally not identifiable.
\end{proposition}
Essentially, \cref{lemma:unidentifiable} follows because the data only identifies the marginals $p_{10}+p_{11},\,p_{01}+p_{11}$ while $\tpr_a,\tnr_a$ depend on the joint via $p_{01}$, which can vary even while marginals are fixed. Since this can vary independently across values of $A$, discrepancies are not identifiable either.

\subsection{Identification under Monotonicity}

We next show identifiability if we impose the additional assumption of monotone treatment response.
\begin{assumption}[Monotone treatment response]\label{asm:monotone}
	$Y(1) \geq Y(0)$. (Equivalently, $p_{10}=0$.)
\end{assumption}
\Cref{asm:monotone} says that anti-responders do not exist. In other words, the treatment either does nothing (e.g., an individual would have gotten a job or not gotten a job, regardless of receiving job training) or it benefits the individual (would get a job if and only if receive job training), but it never harms the individual. This assumption is reasonable for positive interventions. 
As \cite{k19-monotonicity} points out, policy learning in this setting is equivalent to the binary classification problem of predicting responder status.

\begin{proposition}\label{prop-id}
Under \cref{asm:monotone},
\begin{equation}\label{eq:id}
\begin{aligned}
\tpr_a&=\frac{ \Eb{\tau   \mid A=a, Z=1} \Prb{Z=1\mid A=a}  }{\Eb{\tau \mid A=a} }, \\
\tnr_a&=\frac{\Eb{ (1 - \tau)  \mid A=a, Z=0} \Prb{Z=0\mid A=a} }{ \Eb{(1-\tau) \mid A=a} }.
\end{aligned}
\end{equation}
\end{proposition}
Since the quantities on the right hand sides in \cref{eq:id} are in terms of identified quantities (functions of the distribution of $(X,A,T,Y)$), this proves identifiability. Given a sample and an estimate of $\tau$, it also provides a simple recipe for estimation by replacing each average or probability by a sample version, since both $A$ and $Z$ are discrete.

Thus, \cref{prop-id} provides a novel means of assessing disparate impact of personalized interventions under monotone response. This is relevant because monotonicity is a defensible assumption in the case of many interventions that bestow an additional benefit, good, or resource, such as the ones mentioned in \cref{sec:intro}. Nonetheless, 
the validity of \cref{asm:monotone} is itself not identifiable. Therefore, should it fail even slightly, it is not immediately clear whether these disparity estimates can be relied upon. We therefore next study a sensitivity analysis by means of constructing partial identification bounds for $\tpr_a,\tnr_a$.

\section{Partial Identification Bounds for Sensitivity\break Analysis}\label{sec-pi-bounds}

We next study the partial identification of disparate impact metrics when \cref{asm:monotone} fails, i.e., $p_{10}\neq0$. 
We first state a more general version of \cref{prop-id}.
For any $\eta=\eta(X,A)$, let
\begin{align*}\rho^{\tp}_a(\eta) &\coloneqq  \frac{ \Eb{ \tau + \eta \mid {A=a, Z=1} } \Prb{Z=1\mid A=a} }{ \Eb{\tau+ \eta \mid A=a} },\\
\rho^{\tn}_a(\eta) &\coloneqq  \frac{ \Eb{ 1-(\tau + \eta) \mid {A=a,Z=0} } \Prb{Z=0\mid A=a} }{ \Eb{1-(\tau + \eta) \mid A=a} }.
\end{align*}
\begin{proposition}\label{prop-id2}
$\tpr_a=\rho^{\tp}_a(p_{10}),\,\tnr_a=\rho^{\tn}_a(p_{10})$.
\end{proposition}

Since the anti-responder probability $p_{10}$ is unknown, 
we cannot use \cref{prop-id2} to identify $\tp_a, \tn_a$.
We instead use \cref{prop-id2} to compute bounds on them
by restricting $p_{10}$ to be in an uncertainty set. 
Formally, given an uncertainty set $\mathcal U$ for $p_{10}$ (i.e., a set of functions of $x,a$), we define the simultaneous identification region of the TPR and TNR for all groups $a\in\mathcal A$ as:
$$
\Theta=\braces{\prns{\rho^{\tp}_a(\eta),\rho^{\tn}_a(\eta)}_{a\in\mathcal A}\;:\:\eta\in\mathcal U}\subseteq\R{2\times\abs{\mathcal A}}.
$$
For brevity, we will let $\rho_a(\eta)=\prns{\rho^{\tp}_a(\eta),\rho^{\tn}_a(\eta)}$ and $\rho(\eta)=(\rho_a(\eta))_{a\in\mathcal A}$.

The set $\Theta$ describes all possible simultaneous values of the group-conditional true positive and true negative rates. 
As long as $\forall\eta\in\mathcal U$ we have $0\leq\eta(X,A)\leq \min\prns{\Prb{Y=1\mid T=0,X,A},\Prb{Y=0\mid T=1,X,A}}$ (which is identified from the data) by \cref{prop-id2} this set is necessarily sharp \citep{manski2003partial} given only the restriction that $p_{10}\in\mathcal U$.
(In particular, this bound on $\eta$ can be achieved by just point-wise clipping $\mathcal U$ with this identifiable bound as necessary.)
That is, given a joint on $(X,A,T,Y)$, 
on the one hand, every $\rho\in\Theta$ is realized
by some full joint distribution on $(X,A,T,Y(0),Y(1))$ with $p_{10}\in\mathcal U$, and on the other hand, every such joint gives rise to a $\rho\in\Theta$.
In other words, $\Theta$ is an \emph{exact} characterization of the in-fact possible simultaneous values of the group-conditional TPRs and TNRs.

Therefore, if, for example, we are interested in the minimal and maximal possible values for the true (unknown) TPR discrepancy between groups $a$ and $b$, we should seek to compute
$
\inf_{\rho\in\Theta}\ 
\rho^{\tp}_a-\rho^{\tp}_b$ and $\sup_{\rho\in\Theta}\ 
\rho^{\tp}_a-\rho^{\tp}_b.
$
More generally, for any $\mu\in\R{2\times\abs{\mathcal A}}$, we may wish to compute 
\begin{equation}\label{eq:supportfn}h_{ \Theta } ({\mu}) \coloneqq
  \sup_{\rho \in \Theta}  {\mu}^\top \rho . \end{equation}
Note that this, for example, covers the above example since for any $\mu$ we can also take $-\mu$. The function $h_{ \Theta }$ is known as the \emph{support function} of ${ \Theta }$ \citep{rockafellar2015convex}. Not only does the support function provide the maximal and minimal contrasts in a set, it also exactly characterizes its convex hull. That is, $\op{Conv}\prns{ \Theta }=\braces{\rho:\mu^\top\rho\leq h_{\Theta}(\mu)\ \forall\mu}$.
So computing $h_{ \Theta }$ allows us to compute $\op{Conv}\prns{ \Theta }$.

Our next result gives an explicit program to compute the support function when $\mathcal U$ has a product form
of within-group uncertainty sets:
\begin{equation}\label{eq:productU}\mathcal U=\braces{\eta:\eta(\;\cdot\;,a)\in\mathcal{U}_{a}\ \forall a\in\mathcal A},\end{equation}
which leads to $\Theta=\prod_{a\in\mathcal A}\Theta_a$ where $\Theta_a=\braces{\rho_a(\eta_a):\eta_a\in\mathcal U_a}$.
\begin{proposition}\label{prop-supp-fn-rep}
Let $r_a^z \coloneqq \Prb{Z=z\mid A=a}$ and $\tau_a^z\coloneqq\Eb{\tau\mid A=a,Z=z}$. 
Suppose $\mathcal U$ is as in \eqref{eq:productU}.
Then \cref{eq:supportfn} can be reformulated as: %
	\begin{align*}
h_{\Theta} ({\mu})=~&\omit{\rlap{$\sum_{a\in\mathcal A}h_{\Theta_a} ({\mu_a})$}}\\
h_{\Theta_a} ({\mu_a})=~&\sup_{\omega_a, t_a} && %
  \mu_a^{\tp}  r_a^1 \;%
  \prns{t_a\tau_a^1+\Eb{\omega_a(X)\mid {A =a, Z=1}}}
	\\&&&+  \frac{\mu_a^{\tn} r_a^0 }{{t_a}-1} (t_a \;
  \prns{1-\tau_a^0}
  + \Eb{\omega_a(X) \mid {A=a, Z=0}} ) 
  \\
&\quad\mathrm{s.t. }	&& 
\omega_a(\cdot)\in t_a\;\mathcal U_a,
~~ t_a\prns{r_a^0\tau_a^0+r_a^1\tau_a^1}+\Eb{\omega_a \mid A=a}=1.%
	\end{align*}
\end{proposition}
For a fixed value of $t_a$, the above program is a linear program, given that $\mathcal U_a$ is linearly representable.
Therefore a solution may be found by grid search on the univariate $t_a$.
Moreover, if $\mu_a^{\tp}=0$ or $\mu_a^{\tn}=0$, the above remains a linear program even with $t_a$ as a variable \citep{charnes1962programming}.
With this, we are able to express group-level disparities through assessing the support function at specific contrast vectors $\mu$.

\subsection{Partial Identification under Relaxed Monotone Treatment Response}

We next consider the implications of the above for the following relaxation of the monotone treatment response assumption:
\begin{assumption}[$B$-relaxed monotone treatment response]\label{asm:monotone-B}
   $p_{10}\leq B$.
\end{assumption}
Note that \cref{asm:monotone-B} with $B=0$ recovers \cref{asm:monotone}
and \cref{asm:monotone-B} with $B=1$ is a vacuous assumption.
In between these two extremes we can consider milder or stronger violations of monotone response and the partial identification bounds they corresponds to. This provides us with a means of sensitivity analysis of the disparities we measure, recognizing that monotone response may not hold exactly and that disparities may not be exactly identifiable.
For the rest of the paper, we focus solely on partial identification under \cref{asm:monotone-B}.
Note that \cref{asm:monotone-B} corresponds exactly to the uncertainty set $\mathcal U_B=\braces{\eta:0\leq\eta(X,A)\leq \min\prns{B,\Prb{Y=1\mid T=0,X,A},\Prb{Y=0\mid T=1,X,A}}}$.\break We define $\Theta_B=\prod_{a\in\mathcal A}\Theta_{B,a}$ to be the corresponding identification region.

Under \cref{asm:monotone-B}, our bounds take on a particularly simple form. 
Let \begin{align*}\mathcal B_a^z(B)=\E\bigl[\min\bigl(B,\,&\Prb{Y=1\mid T=0,X,A},\\&\Prb{Y=0\mid T=1,X,A}\bigr)\mid A=a,Z=z\bigr]\end{align*} and
define
\begin{align*}
\overline{\rho}^{\tp}_a(B) &=  \frac{ (\tau_a^1+\mathcal B_a^1(B)) r_a^1 }{ 
\tau_a^0 r_a^0+(\tau_a^1+\mathcal B_a^1(B)) r_a^1},\\
\underline{\rho}^{\tp}_a(B) &=\frac{ \tau_a^1 r_a^1 }{ 
(\tau_a^0+\mathcal B_a^0(B)) r_a^0+\tau_a^1 r_a^1},\\
\overline{\rho}^{\tn}_a(B) &=  
\frac{ (1-\tau_a^0) r_a^0 }{ 
(1-\tau_a^0) r_a^0+(1-\tau_a^1-\mathcal B_a^1(B)) r_a^1}
,\\
\underline{\rho}^{\tn}_a(B) &=  
\frac{ (1-\tau_a^0-\mathcal B_a^0(B)) r_a^0 }{ 
(1-\tau_a^0-\mathcal B_a^0(B)) r_a^0+(1-\tau_a^1) r_a^1}
. 
\end{align*}

\begin{proposition}\label{prop-flp-1class}
Suppose \cref{asm:monotone-B} holds. Then $[\underline{\rho}^{\tp}_a(B),\overline{\rho}^{\tp}_a(B)]$ and $[\underline{\rho}^{\tn}_a(B),\overline{\rho}^{\tn}_a(B)]$ are the sharp identification intervals for $\tp_a$ and $\tn_a$, respectively. Moreover,
$(\underline{\rho}^{\tp}_a(B),\underline{\rho}^{\tn}_a(B))\in\Theta_{B,a}$ and
$(\overline{\rho}^{\tp}_a(B),\overline{\rho}^{\tn}_a(B))\in\Theta_{B,a}$, i.e., the two extremes are simultaneously achievable.
\end{proposition}

\section{Partial Identification of Group Disparities and ROC and xROC Curves}\label{sec-group-disparities}
We discuss diagnostics
to summarize possible impact disparities across a {range} of possible policies.  %

\paragraph{TPR and TNR disparity.} 
Discrepancies in model errors (TPR or TNR) are of interest when auditing classification performance on different groups with a given, fixed policy $Z$. Under \cref{asm:monotone}, they are identified by \cref{prop-id}. 
Under violations of \cref{asm:monotone}, we can consider their partial identification bounds. If the \emph{minimal} disparity remains nonzero, that provides strong evidence of disparity. Similarly, if the \emph{maximal} disparity is large, a responsible decision maker should be concerned about the possibility of a disparity.

Under \cref{asm:monotone-B}, \cref{prop-flp-1class} provides that the sharp identification intervals of $\tpr_a-\tpr_b$ and $\tnr_a-\tnr_b$ are, respectively, given by
\begin{equation}\label{eq-tpr-tnr-disp}\begin{aligned}
&[\underline \rho_a^{\tp}(B)-\overline \rho_b^{\tp}(B),\ \overline \rho_a^{\tp}(B)-\underline \rho_b^{\tp}(B)],\\&[\underline \rho_a^{\tn}(B)-\overline \rho_b^{\tn}(B),\ \overline \rho_a^{\tn}(B)-\underline \rho_b^{\tn}(B)].
\end{aligned}\end{equation}
Given effect scores $\tau$, we can then use this to plot \emph{disparity curves} by plotting the endpoints of \cref{eq-tpr-tnr-disp} for policies $Z=\mathbb{I}[\tau \geq \theta]$ for varying thresholds $\theta$.

\paragraph{Robust ROC Curves} 
We first define the analogous \textit{group-conditional} \textit{ROC curve} corresponding to a CATE function $\tau$. These are the parametric curves traced out by the pairs $(1-\tnr_a,\tpr_a)$ of policies that threshold the CATE for varying thresholds. To make explicit that we are now computing metrics for different policies, we use the notation ${\rho}(\eta; \tau \geq \theta)$ to refer to the metrics of the policy $Z=\indic{\tau\geq\theta}$.
Under \cref{asm:monotone}, \cref{prop-id} provides point identification of the group-conditional ROC curve:
\begin{align*} &{\op{ROC}}_a(\tau) \coloneqq \{ (1-	{\rho}_a^{\tn}(0; \tau \geq \theta),{\rho}_a^{\tp}(0; \tau \geq \theta)  ) : \theta \in \Rl \}
\end{align*}
When \cref{asm:monotone} fails, we cannot point identify $\tpr_a,\tnr_a$ and correspondingly we cannot identify ${\op{ROC}}_a(\tau)$.
We instead define the \emph{robust ROC} curve as the union of all partially identified ROC curves. Specifically:
\begin{align*} &{\Theta^{\op{ROC}}_a}(\tau) \coloneqq \{ (1-  {\rho}^{\tn}_a(\eta_a; \tau \geq \theta),{\rho}^{\tp}_a(\eta_a; \tau \geq \theta)  ) \colon \theta \in \Rl, \eta_a \in \mathcal{U}_a \}.
\end{align*}
Plotted, this set provides a visual representation of the region that the true ROC curve can lie in. We next prove that under \cref{asm:monotone-B}, we can easily compute this set as the area between two curves.
\begin{proposition}
Let $\mathcal{U}=\mathcal{U}_{B}$. Then ${\Theta^{\op{ROC}}_a}(\tau)$ is given as the area between the two parametric curves 
$\underline{\op{ROC}}_a(\tau)\coloneqq\{ (1-  {\underline\rho}_a^{\tn}(B; \tau \geq \theta),{\underline\rho}_a^{\tp}(B; \tau \geq \theta)  ) \colon \theta \in \Rl \}$
and 
$\overline{\op{ROC}}_a(\tau)\coloneqq\{ (1-  {\overline\rho}_a^{\tn}(B; \tau \geq \theta),{\overline\rho}_a^{\tp}(B; \tau \geq \theta)  ) \colon \theta \in \Rl \}$.
\end{proposition}
This follows because the extremes are simultaneously achievable as noted in \cref{prop-flp-1class}. We highlight, however, that the lower (resp., upper) ROC curve may not be simultaneously realizable as an ROC curve of any single policy.

\paragraph{Robust xROC Curves} 
Comparison of group-conditional ROC curves may not necessarily show impact disparities as, even in standard classification settings ROC curves can overlap despite disparate impacts \citep{kz-19,hardt2016equality}.
At the same time, comparing disparities for fixed policies $Z$ with fixed thresholds may not accurately capture the impact of using $\tau$ for rankings.
\cite{kz-19} develop the $\op{xAUC}$ metric for assessing the \textit{bipartite ranking} quality of risk scores, as well as the analogous notion of a $\op{xROC}$ curve which parametrically plots the TPR of one group vs. the FPR of \textit{another group}, at any fixed threshold. 
This is relevant if effect scores $\tau$ are used for downstream decisions by different facilities with different budget constraints or if the score is intended to be used by a ``human-in-the-loop'' exercising additional judgment, e.g., individual caseworkers as in the encouragement design of \cite{behncke-evaluation}. 

Under \cref{asm:monotone}, we can point identify $\tpr_a,\tnr_a$, so, following \cite{kz-19}, we can define the point-identified xROC curve as
	\begin{align*} &\op{xROC}_{a,b}(\tau)= \{ 	(1-	{\rho}_b^{\tn}(0; \tau \geq \theta),{\rho}_a^{\tp}(0; \tau \geq \theta)  ) :\theta\in\Rl\}.
	\end{align*}
Without \cref{asm:monotone},
we analogously define the \emph{robust xROC} curve as the union of all partially identified xROC curves:
\begin{align*} &\Theta^{\op{xROC}}_{a,b}(\tau)= \{   (1- {\rho}_b^{\tn}(\eta_a; \tau \geq \theta),{\rho}_a^{\tp}(\eta_a; \tau \geq \theta)  ) :\theta\in\Rl,\eta_a\in\mathcal U_a\}.
  \end{align*}
\begin{proposition}
Let $\mathcal{U}=\mathcal{U}_{B}$. Then ${\Theta^{\op{xROC}}_{a,b}}(\tau)$ is given as the area between the two parametric curves 
$\underline{\op{xROC}}_{a,b}(\tau)\coloneqq\{ (1-  {\underline\rho}_b^{\tn}(B; \tau \geq \theta),{\underline\rho}_a^{\tp}(B; \tau \geq \theta)  ) \colon \theta \in \Rl \}$
and 
$\overline{\op{xROC}}_{a,b}(\tau)\coloneqq\{ (1-  {\overline\rho}_b^{\tn}(B; \tau \geq \theta),{\overline\rho}_a^{\tp}(B; \tau \geq \theta)  ) \colon \theta \in \Rl \}$.
\end{proposition}
This follows because $\mathcal U_B$ takes the form of a product set over $a\in\mathcal A$.

\section{Case Study: Personalized Job Training\break (Behaghel et al.)}\label{sec-behaghel}

\begin{figure*}[t!]\centering
				\includegraphics[width=0.4\linewidth]{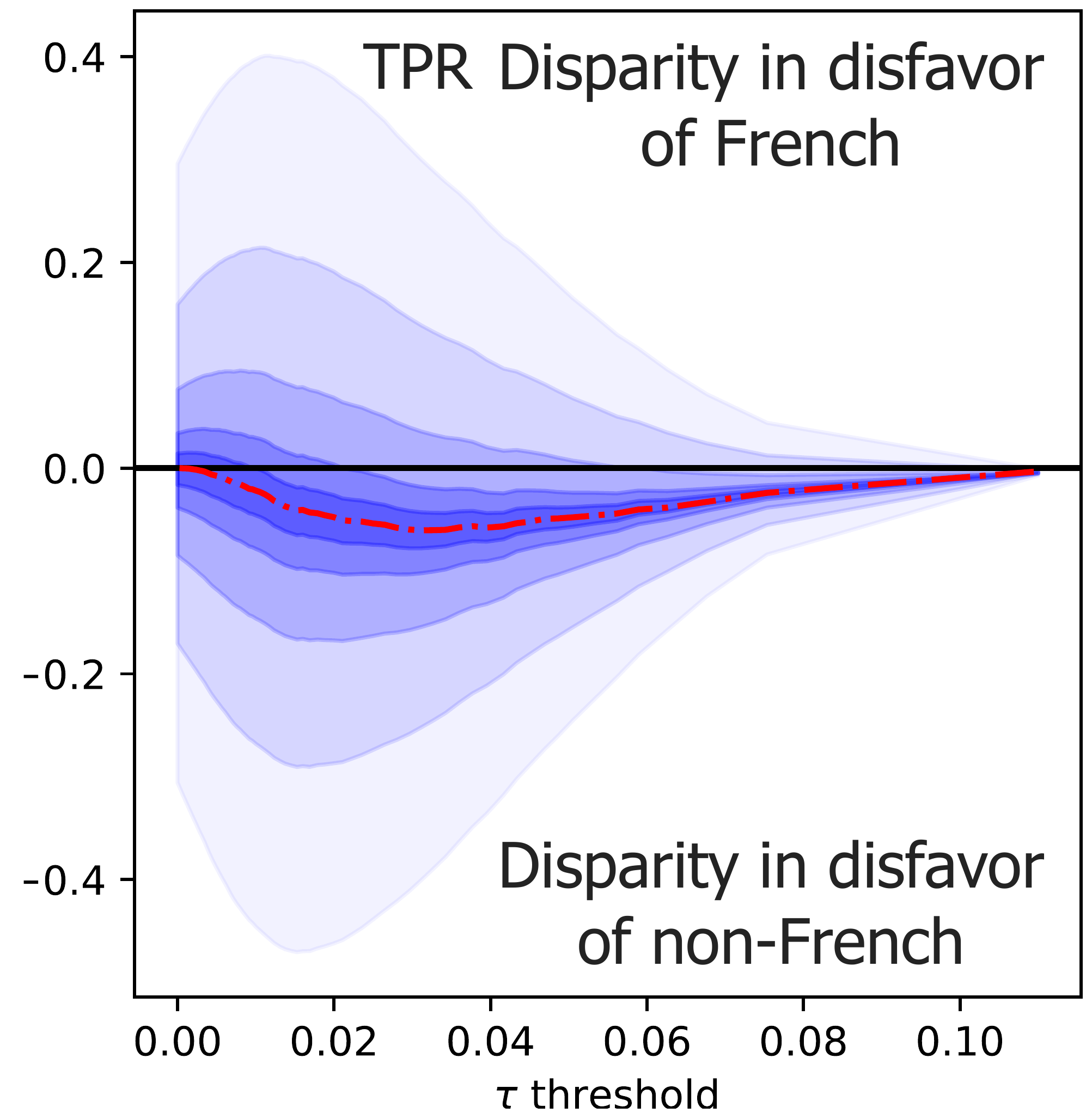}\hspace{0.1\linewidth}\includegraphics[width=0.4\linewidth]{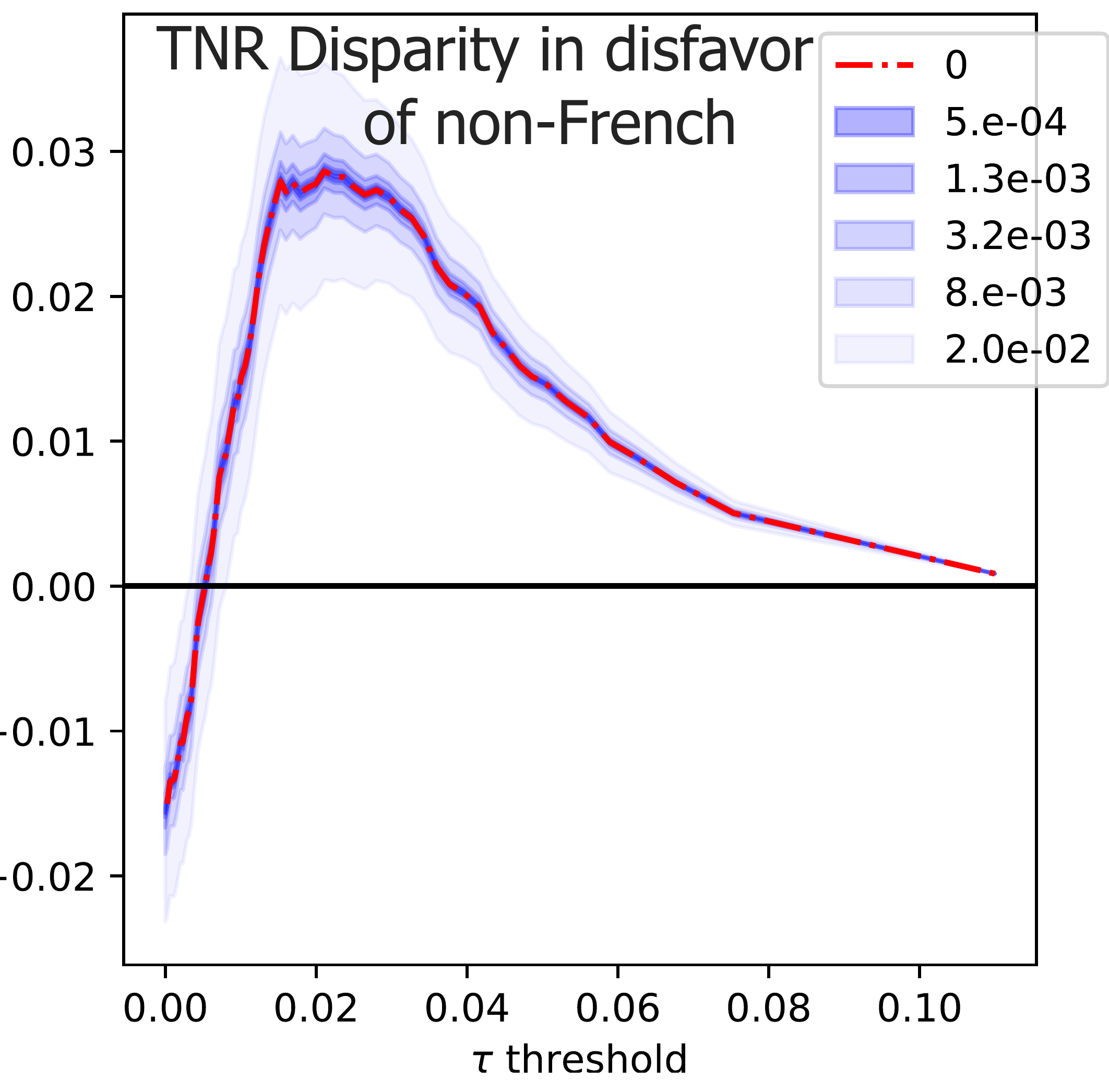}
		\includegraphics[width=0.4\linewidth]{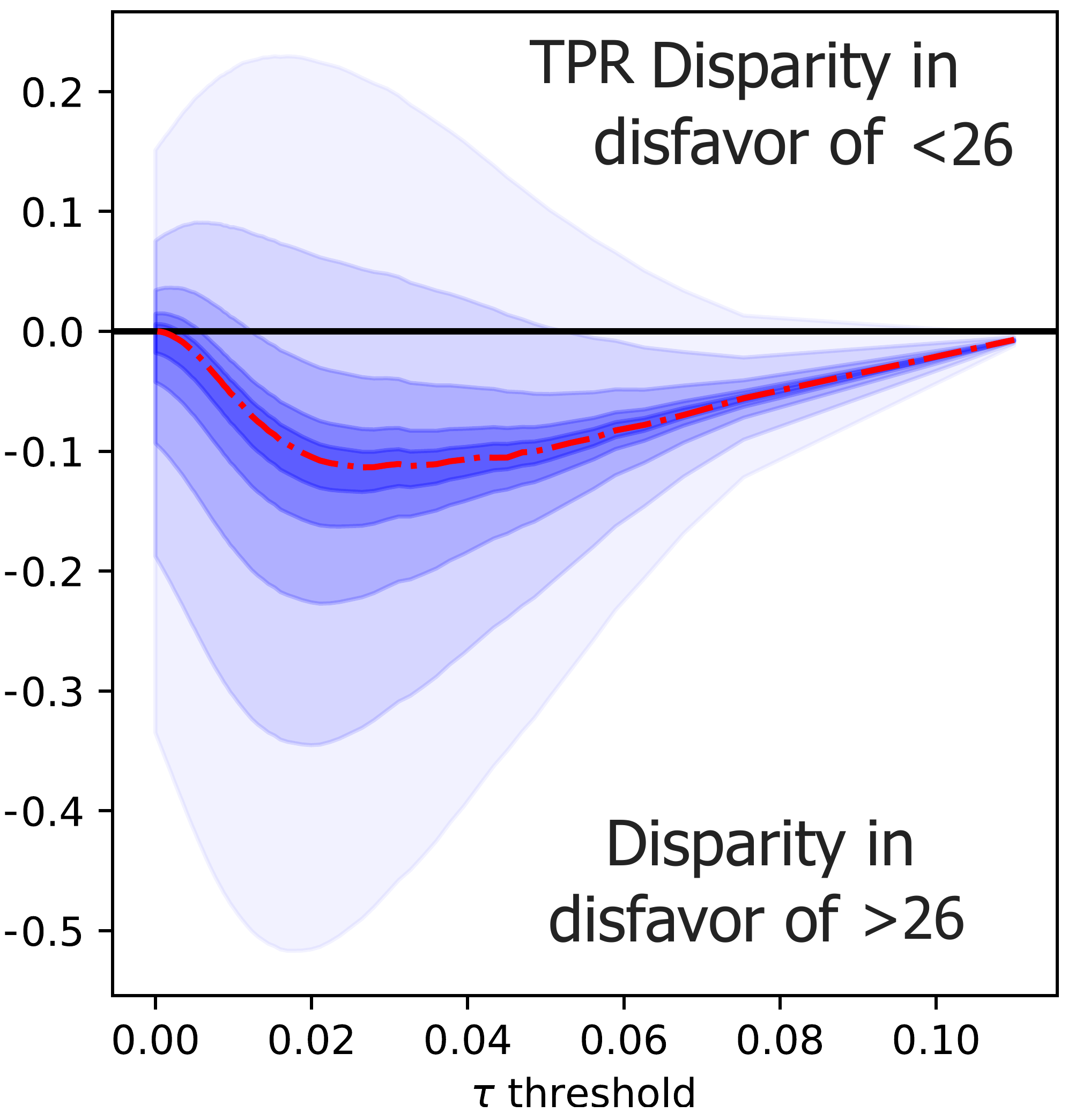}\hspace{0.1\linewidth}\includegraphics[width=0.4\linewidth]{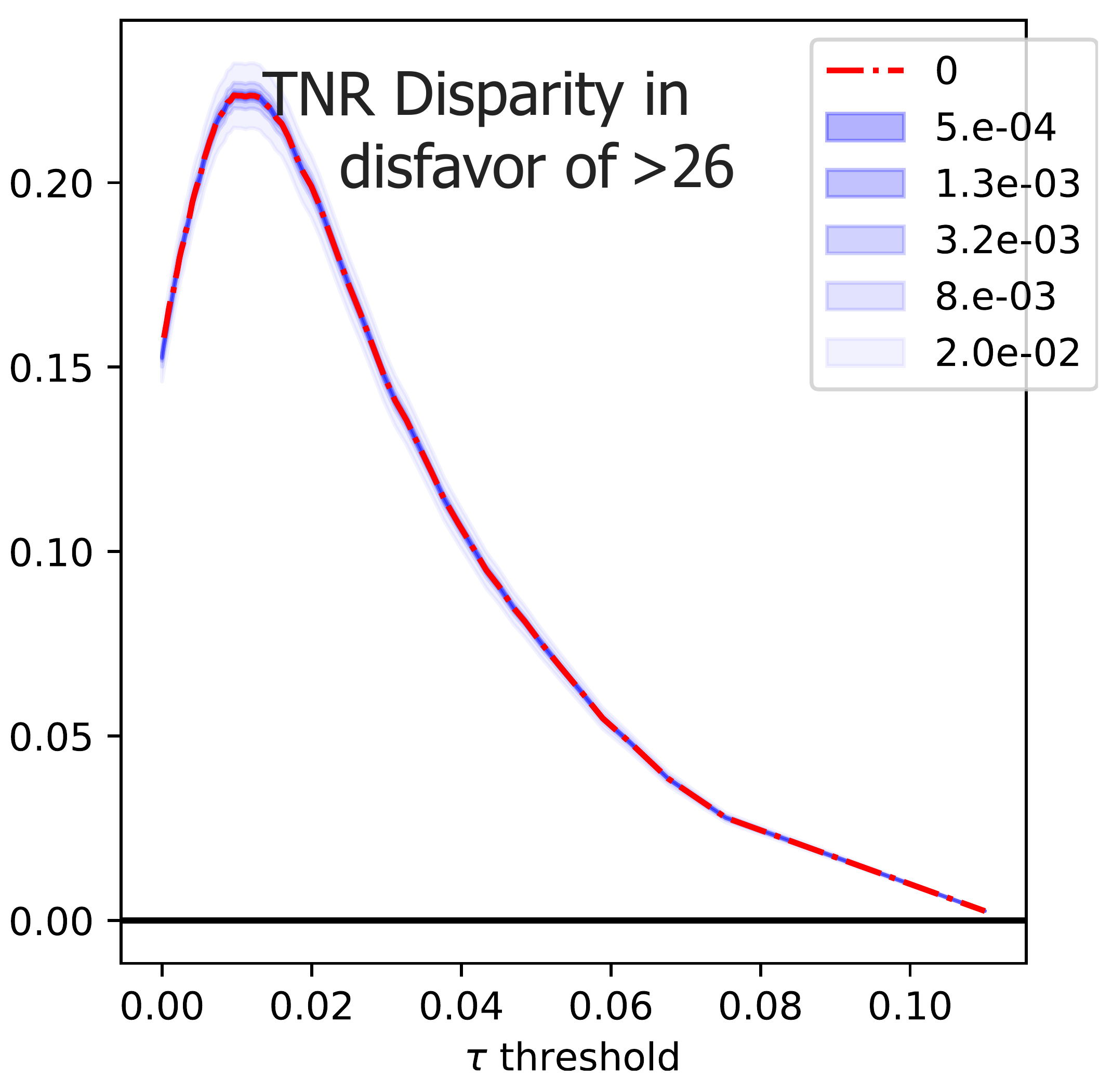}
		\caption{TPR and TNR disparity curves and bounds on French job training dataset (\cref{eq-tpr-tnr-disp}) }\label{fig-disparity-curves}
\end{figure*}

We consider a case study from a three-armed large randomized controlled trial that randomly assigned job-seekers in France to a control-group, a job training program managed by a public vendor, and an out-sourced program managed by a private vendor \cite{behaghel-2014}. While the original experiment was interested in 
the design of contracts for program service delivery, we consider a task of heterogeneous causal effect estimation, motivated by interest in personalizing different types of counseling or active labor market programs that would be beneficial for the individual. Recent work in policy learning has also considered personalized job training assignment \cite{wager17,kt15} and suggested excluding sensitive attributes from the input to the decision rule for fairness considerations, but without consideration
of fairness in the causal effect estimation itself and how significant impact disparities may still remain after excising sensitive attributes because of it. 

We focus on the public program vs. control arm, which enrolled about 7950 participants in total, with $n_1 = 3385$ participants in the public program. The treatment arm, $T=1$, corresponds to assignment to the public program. The original analysis suggests a small but statistically significant positive treatment effect of the public program, with an ATE of $0.023$. We omit further details on the data processing to \cref{apx-crepon-data}. We consider the group indicators: 
\textit{nationality} ($0,1$ denoting French nationals vs. non-French, respectively), \textit{gender} (denoting woman vs. non-woman), and \textit{age} (below the age of 26 vs. above).  (Figures for gender appear in \cref{apx-crepon-data}.)

In \cref{fig-disparity-curves}, we plot the identified ``disparity curves'' of \cref{eq-tpr-tnr-disp} corresponding to the maximal and minimal sensitivity bounds on TPR and TNR disparity between groups. Levels of shading correspond to different values of $B$, with color legend at right. We learn $\tau$ by the Generalized Random Forests method of \cite{wager2017estimation,athey2019generalized} and use sample splitting, learning $\tau$ on half the data and using our methods to assess bounds on $\rho^{\tpr}, \rho^{\tnr}$ and other quantities with out-of-sample estimates on the other half of the data. We bootstrap over 50 sampled splits and average disparity curves to reduce sample uncertainty.

In general, the small probability of being a responder leads to increased sensitivity of TPR estimates (wide identification bands). The curves and sensitivity bounds suggest that with respect to nationality and gender, there is small or no disparity in true positive rates but 
the true negative rates for nationality, gender, and age may differ significantly across groups, such that non-women would have a higher chance of being bestowed job-training benefits when they are in fact not responders. However, TPR disparity by age appears to hold with as much as -0.1 difference, with older actually-responding individuals being less likely to be given job training than younger individuals. Overall, this suggests that differences in heterogeneous treatment effects across age categories could lead to significant adverse impact on older individuals. 

\begin{figure*}[t!]\centering
\includegraphics[width=0.4\linewidth]{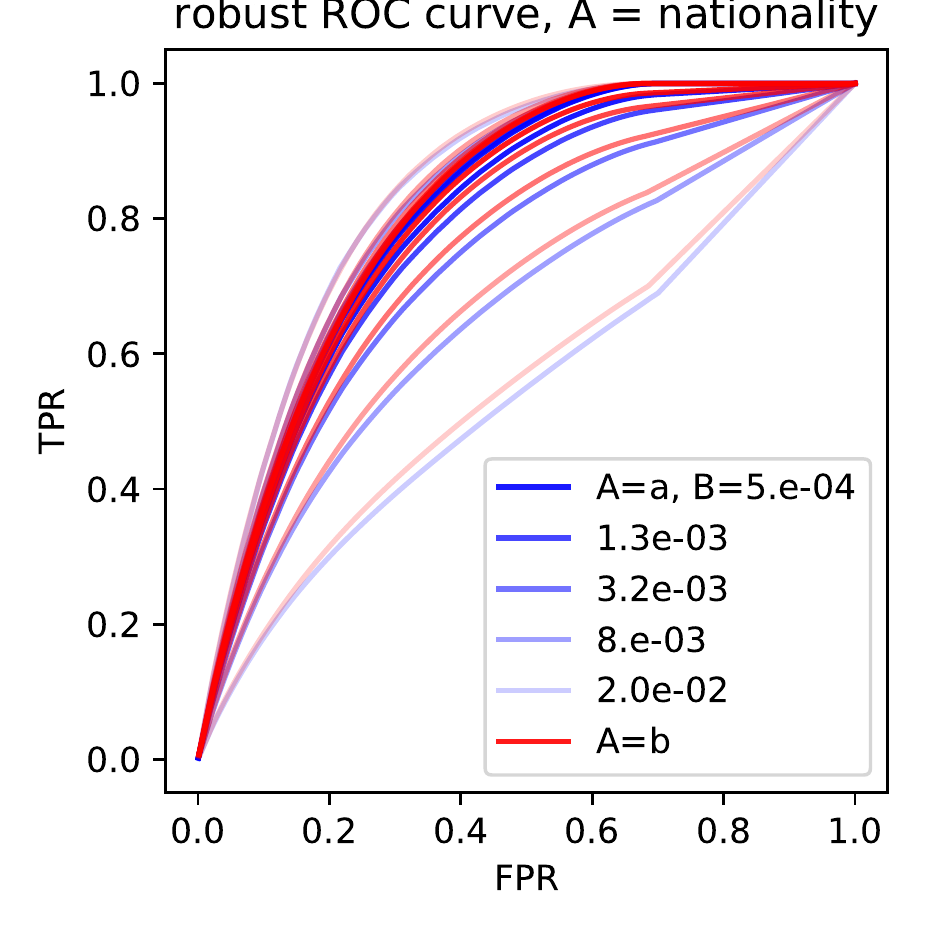}\hspace{0.1\linewidth}\includegraphics[width=0.4\linewidth]{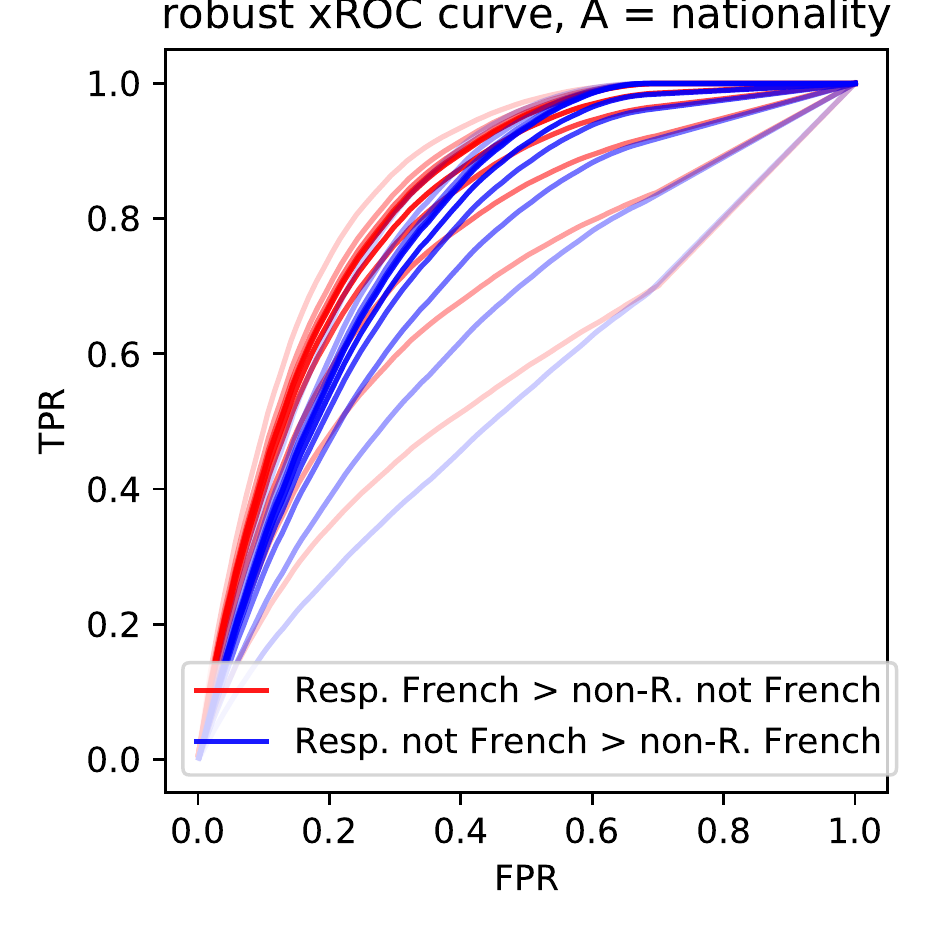}
\includegraphics[width=0.4\linewidth]{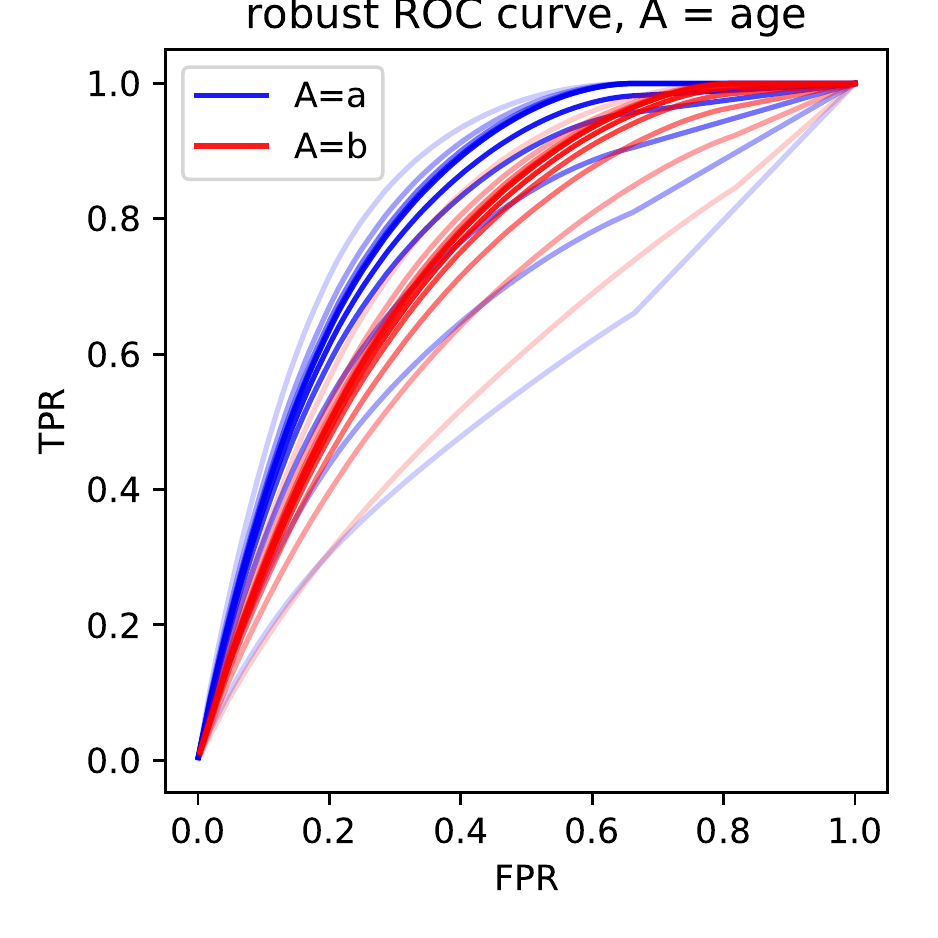}\hspace{0.1\linewidth}\includegraphics[width=0.4\linewidth]{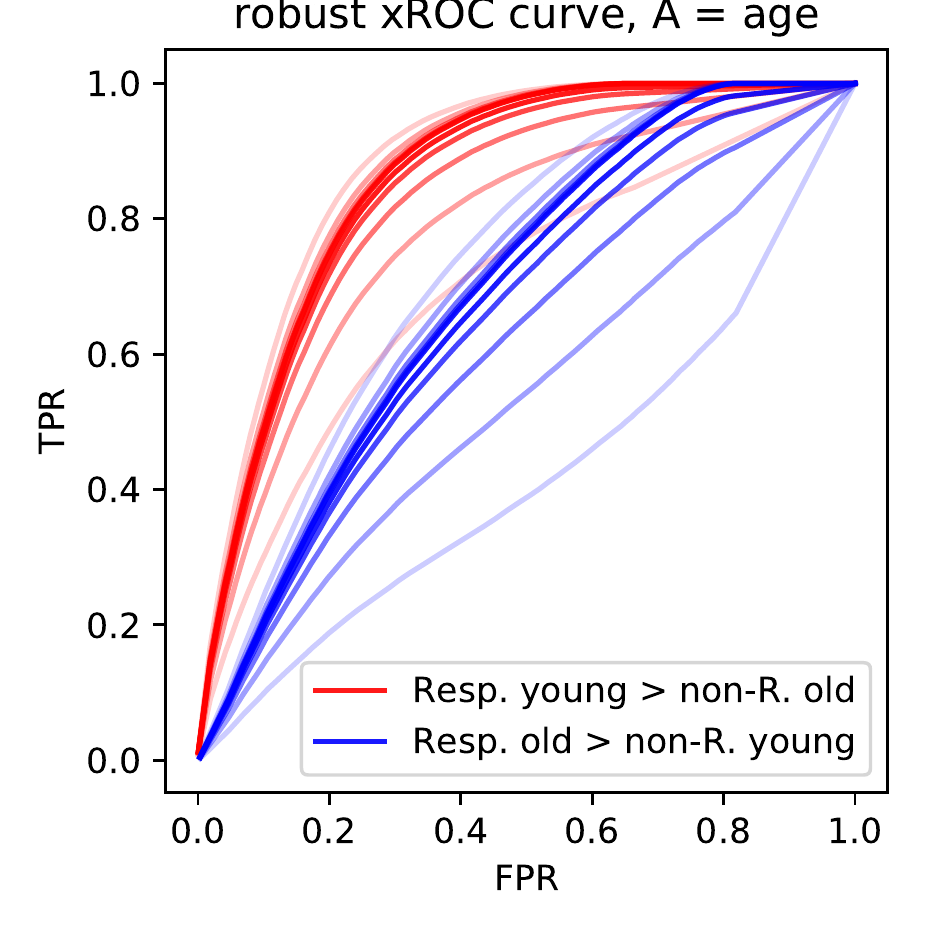}
\caption{ROC and xROC for $A=$ nationality, age on French job training dataset %
}
  \label{fig-xroc}
\end{figure*}

This is similarly reflected in the robust ROC, xROC curves (\cref{fig-xroc}). Despite possibly small differences in ROCs, the xROCs indicate strong disparities: the sensitivity analysis suggests that the likelihood of ranking a non-responding young individual above a responding old individual (xAUC \citep{kz-19}) is clearly larger than the symmetric error, meaning that older individuals who benefit from the treatment may be disproportionately shut out of it as seats are instead given to non-responding younger individuals.

\section{Discussion and Conclusion}\label{sec-discussion} 
We presented identification results and bounds for assessing disparate model errors of causal-effect maximizing treatment policies, which can lead disparities in access to those who stand to benefit from treatment across groups.
Whether this is ``unfair'' would naturally rely on one's normative assumptions. One such is ``claims across outcomes,''
that individuals have a claim to the public intervention if they stand to benefit, which can be understood within \cite{adler}'s axiomatic justification of fair distribution. 
 There may also be other justice-based considerations, e.g. minimax fairness. We discuss this more extensively in \cref{apx-substantive}. 

With the new ability to \textit{assess} disparities using our results, a second natural question is whether these disparities warrant adjustment, which is easy to do given our tools combined with the approach of \citep{hardt2016equality}. This question again is dependent both on one's viewpoint and ultimately on the problem context, and we discuss it further in \cref{apx-substantive}.
Regardless of normative viewpoints, auditing allocative disparities that would arise from the implementation of a personalized rule must be a crucial step of a responsible and convincing program evaluation.
We presented fundamental identification limits to such assessments but provided sensitivity analyses that can support reliable auditing.

\bibliography{beyondfairness,sensitivity}
\bibliographystyle{abbrvnat}

\clearpage
\appendix
\section{Omitted proofs}

\begin{proof}[Proof of \cref{lemma:unidentifiable}] To prove this we exhibit a simple example satisfying ignorability where both $\tpr_a,\tnr_a$ and differences therein varies while the joint distribution of $(X,A,T,Y)$ does not. 

Let $\mathcal X=\{0,1\}$, $Z=\indic{X=1}$,
$\Prb{T=t,X=x\mid A=a}=\frac14$, $\Prb{A=a}=1/\abs{\mathcal A}$.
To specify a joint distribution of $(X,A,T,Z,Y(1),Y(0))$ that satisfies ignorable treatment, it only remains to specify $p_{ij}$.

Note that in this case
$$\tpr_a=\frac{p_{01}(1,a)}{p_{01}(0,a)+p_{01}(1,a)},~
\tpr_a=\frac{1-p_{01}(0,a)}{2-p_{01}(0,a)-p_{01}(1,a)}.$$

The result follows by noting that where the corresponding joint distribution of $(X,A,T,Z,Y)$ is completely specified by $p_{01}+p_{11},\ p_{10}+p_{11}$, while $p_{01}$ could vary as long as these sums are neither 0 nor 1. Since we can vary this independently across values of $A$, differences are not identifiable either.
\end{proof}

\begin{proof}[Proof of \cref{prop-id}]
  \begin{align*}
  & \mathbb P(Z=1\mid A=a,Y(1)>Y(0)) \\
  =& \frac{\mathbb P(Y(1)>Y(0)\mid A =a, Z=1) \mathbb P(Z=1\mid A=a) }{ \mathbb P(Y(1)>Y(0)\mid A=a) }%
  \\
  =& \frac{ \mathbb E [  \E[ Y(1)=1 \mid \substack{T=1\\A =a,X=x} ] - \E[ Y(0)=1 \mid \substack{T=0\\A =a,X=x }]\mid \substack{Z=1\\A=a} ] \mathbb P(Z=1\mid A=a) }{\mathbb E [ \mathbb P(Y(1)=1 \mid \substack{T=1\\A =a,X=x}) - \mathbb P(Y(0)=1 \mid \substack{T=0\\A =a,X=x })\mid A=a]} \;%
  \\
  =& \frac{\mathbb E [ \tau(X,A) \mid \substack{Z=1 \\A =a}] \mathbb P(Z=1\mid A=a) }{ \E[ \tau(X,A) \mid A =a ]  } \;%
  \end{align*}
  where the first equality holds by Bayes' rule, the second by iterating expectations on $X$ and \cref{asm:monotone}, and the third by unconfoundedness and consistency of potential outcomes. The proof for identification of $\tn$ is identical for the quantity $\mathbb P(Z=0\mid A=a,Y(1)\leq Y(0)) $.
\end{proof}

\begin{proof}[Proof of \cref{prop-id2}]
Recalling that CATE identifies, under violations of \cref{asm:monotone} $$\tau=\Efb{Y(1)-Y(0)\mid X,A} =p_{01}-p_{10},$$
  \begin{align*}
=\;& \frac{\mathbb E [ \tau + \eta \mid \substack{Z=1 \\A =a}] \mathbb P(Z=1\mid A=a) }{ \E[ \tau + \eta \mid A =a ]  } =\frac{ (p_{01}-p_{10} + p_{10}) \mathbb P(Z=1\mid A=a) }{\mathbb E [ (p_{01}-p_{10} + p_{10})\mid A=a]} \;\\
=\;& \mathbb P(Z=1\mid A=a,Y(1)>Y(0))
\end{align*}
\end{proof}

\begin{proof}[Proof of \cref{prop-supp-fn-rep}]\label{proof-supp-fn-rep}
	The support function evaluated at $\mu$ is: 
	\begin{align*}
	\max_{\eta} & \sum_{a \in \mathcal{A}} \mu^{\tp}_a  \frac{ \Eb{ \tau + \eta \mid {A=a, Z=1} } r_a^1 }{ \Eb{\tau+ \eta \mid A=a} } + \mu^{\tn}_a \frac{ \Eb{ 1-(\tau + \eta) \mid {A=a,Z=0} } r^a_0 }{ \Eb{1-(\tau + \eta) \mid A=a} } \\
	\mathrm{s.t. }	\;\;& 0 \leq  \eta(x,a) \leq  \mathcal B_a^z(B), \quad \forall x \in \mathcal{X}, \; \forall a \in \mathcal{A}
	\end{align*}
	We apply the Charnes-Cooper transformation \cite{charnes1962programming}with the bijection $t_a = \frac{1}{\Eb{\tau + \eta \mid A =a } },\; \omega_a = \eta t_a$. The denominator of the second term under this bijection is equivalently 
	$$  \Eb{1-(\tau + \eta) \mid A=a}  =  1 - \frac{1}{t_a}$$
	such that we can rewrite the second term as 
	\begin{align*}
	&{\mu_a^{\tn} r_a^0 } \left( \frac{1}{1 - \nicefrac{1}{t_a}} \Eb{ 1-\tau \mid \substack{A=a\\ Z=0} } + \frac{\nicefrac{1}{t_a}}{1 - \nicefrac{1}{t_a}} \Eb{{\omega_a} \mid \substack{A=a\\ Z=0}} \right)  \\&=  \frac{\mu_a^{\tn} r_a^0 }{{t_a}-1} (t \;\Eb{ 1-\tau \mid \substack{A=a\\ Z=0} } + \Eb{\omega_a \mid \substack{A=a\\ Z=0}} ) 
	\end{align*}
	and the objective function overall as: 
		\begin{align*}
	\max_{\eta} & \sum_{a \in \mathcal{A}} (\mu^{\tp}_a r_a^1)   (t_a  \tau_a^1  +\Eb{\omega_a \mid \substack{A=a\\ Z=1}} ) +   \frac{\mu_a^{\tn} r_a^0 }{{t_a}-1} (t_a \;(1-\tau_a^0)+ \Eb{\omega_a \mid \substack{A=a\\ Z=0}} ) 
	\end{align*}
	
	The new constraint set (including the constraint yielding the definition of $t_a$) is: 
	\begin{align*}
	\mathcal{U} = \{ \Eb{\tau t_a + \omega_a \mid A =a } = 1, \;\;
	\omega_a(x,a) \leq t_a \mathcal B_a^z(B), \quad \forall x \in \mathcal{X}, \; \forall a \in \mathcal{A} \} \end{align*}
\end{proof}

\begin{proof}[Proof of \cref{prop-flp-1class}]\label{proof-flp}
We first consider the case of maximizing or minimizing the TPR.

We leverage the invariance in the objective function under the surjection on $\eta(x,a)$ to its marginal expectation over a $Z=z, A=a$ partition.
$$ w(x,a) = \begin{cases}
 \Eb{ \eta \mid Z=1, A=a } & \text{ if }Z = 1 \\
\Eb{ \eta \mid Z=0, A=a } & \text{ if } Z= 0
\end{cases} $$
Therefore we can reparametrize the program as optimizing over coefficients $x,y$ of the optimal solution, $w(x,y)  = x \mathbb{I}[Z = 0] + y \mathbb{I}[Z = 1] $.	Define the fractional objective
	
\begin{align*}g(\alpha,\beta) &=   \frac{ \Eb{ \tau + x \mathbb{I}[Z = 0] +y \mathbb{I}[Z = 1]   \mid {A=a, Z=1} } \Prb{Z=1\mid A=a} }{ \Eb{\tau+ x \mathbb{I}[Z = 0] + y \mathbb{I}[Z = 1]  \mid A=a} } \\
 &= \frac{ (\Eb{ \tau   \mid {A=a, Z=1} }+y) r_a^1  }{ \Eb{\tau \mid A=a} + x r_a^0 + yr_a^1 } 
\end{align*}
	
	First note that without loss of generality that when maximizing, we can set $x = 0$ since this decreases the objective regardless of the value of $y$. We can consider the constrained problem $\max_{y \leq B} h(y)$ where $h(y) = g(0, y)$. Then we have the first and second derivatives, \begin{align*} \frac{\partial h }{\partial y} &= \frac{r_a^1 (\E[\tau\mid A=a] - \E[\tau \mid Z=1, A=a] }{(y r_a^1  +\ \E[\tau \mid A=a]   )^2},\\\frac{\partial^2 h }{\partial \beta^2} &= \frac{(r_a^1)^2 ( \E[\tau \mid A=a] - \E[\tau \mid Z=1,A=a] ) }{(y r_a^1  +\ \E[\tau \mid A=a]     )^3}\end{align*}
	
	By inspection, since $y\geq 0$ we have that $\frac{\partial^2 h }{\partial y^2} \geq 0$ so the function is convex. So when maximizing $h$ on the constraints for $y$, it attains optimal value at the boundary (since $h$ is increasing). When minimizing, note that the derivative is not vanishing anywhere on the constraint set so it suffices to check the endpoints, where the minimum is achieved at $\beta = 0$.

We now consider the case of minimizing or maximizing the TNR. 

Now consider a generic $ f(x) = \frac{a - b x}{c - bx - dy}$ which represents the TNR sensitivity bound with $\omega  = x \mathbb{I}[Z=0] + y \mathbb{I}[Z=1] $, and the constants
	\begin{align*}a &= r_a^0-\E[ \tau \mid Z=0, A=a], \quad	c = 1 - \E[\tau \mid A=a]\\
	b &= r_a^0 , \quad d = r_a^1
	 \end{align*}
	  Without loss of generality we know that we can set $y$ to its upper bound $B$ when maximizing as we are only increasing the objective value; then $c' = c - B r_a^1$. We verify that the second derivative is negative, so that the function is concave: $$ \frac{\partial^2 f}{\partial x^2} = \frac{2 b^2 (a - c')}{(c' -bx)^3} = \frac{2 (r_a^0)^2 ( r_a^0-\E[ \tau \mid Z=0, A=a] - (1 - \E[\tau \mid A=a]-  B r_a^1 ) ) }{(1 - \E[\tau \mid A=a]-  B r_a^1 ) - xr_a^0  } $$
	
	Checking the sign of the numerator simplifies to checking the sign of $$a - c' = (-r_a^1 + \E[ \tau +   B\mid Z=1, A=a])  )$$
	which is negative. The denominator is lower bounded by $1 - \E[\tau \mid A=a] - B$ which is always positive: therefore the problem is concave. The first derivative $ \frac{\partial  f}{\partial x} = \frac{b(a-c')}{(c'-bx)^2}$ is negative on the domain; therefore the maximum is achieved at $x = 0$. Therefore, when maximizing, $\omega = B \mathbb{I}[Z=1] $.

For minimizing the TPR, we take a similar approach: analogously, we can set $y$ to its lower bound without loss of generality. Following the same analysis, the function is still concave $\frac{\partial^2 f}{\partial x^2} = \frac{2 b^2 (a - c')}{(c' -bx)^3}$ since $- r_a^1 -\E[ \tau \mid A=a]<0$ and decreasing with nonzero first-derivative; so the minimum is achieved at $\omega = \mathbb{I}[Z=0]B$.

\end{proof}

	\section{Behaghel et al. Job Training}\label{apx-crepon-data}
	
	We processed the data using replication files available with the AEJ: Applied Economics journal electronic supplement. For the sake of simplicity, we analyze the trial as if it were a randomized controlled trial (without accounting for noncompliance or different randomization probabilities that differ by region). Thus, we consider intention-to-treat effects (as intention to treat is ultimately the policy lever available). We further restricted some covariates, omitting some where personalized allocation based on these covariates seemed unilkely for fairness reasons. The covariates we retain include: length of previous employment, salary, education level, reason for unemployment, region, years of experience at previous job, statistical risk level, job search type (full-time or non-full time), wage target, time of first unemployment spell, job type, and number of children.
	
	An interacted linear model indicates potential heterogeneity of treatment effect with significance on college education, economic layoff, those seeking work due to fixed term contracts or those with previous layoffs. 
	
	\begin{figure}\centering
		\includegraphics[width=0.4\linewidth]{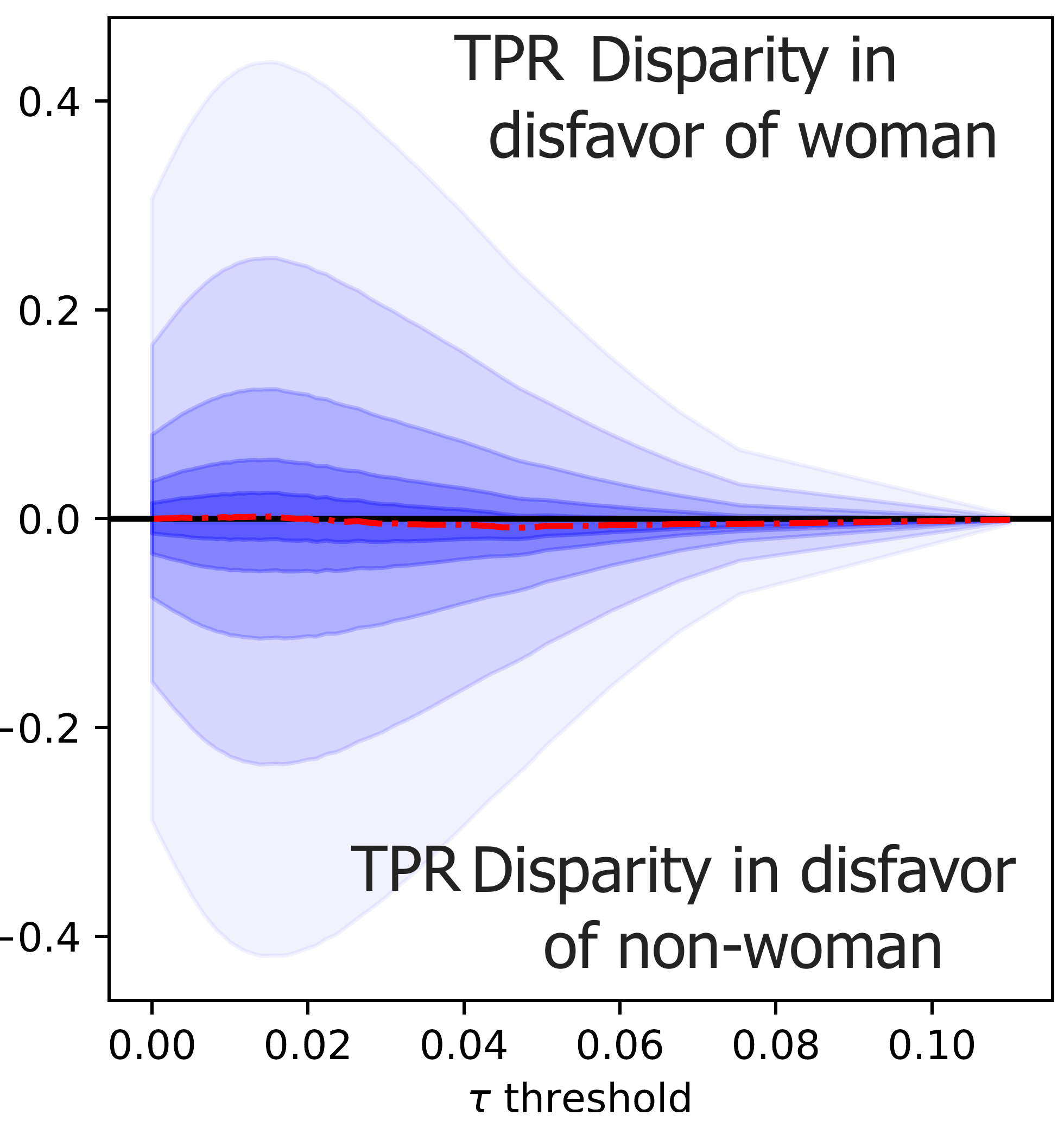}\hspace{0.1\linewidth}\includegraphics[width=0.4\linewidth]{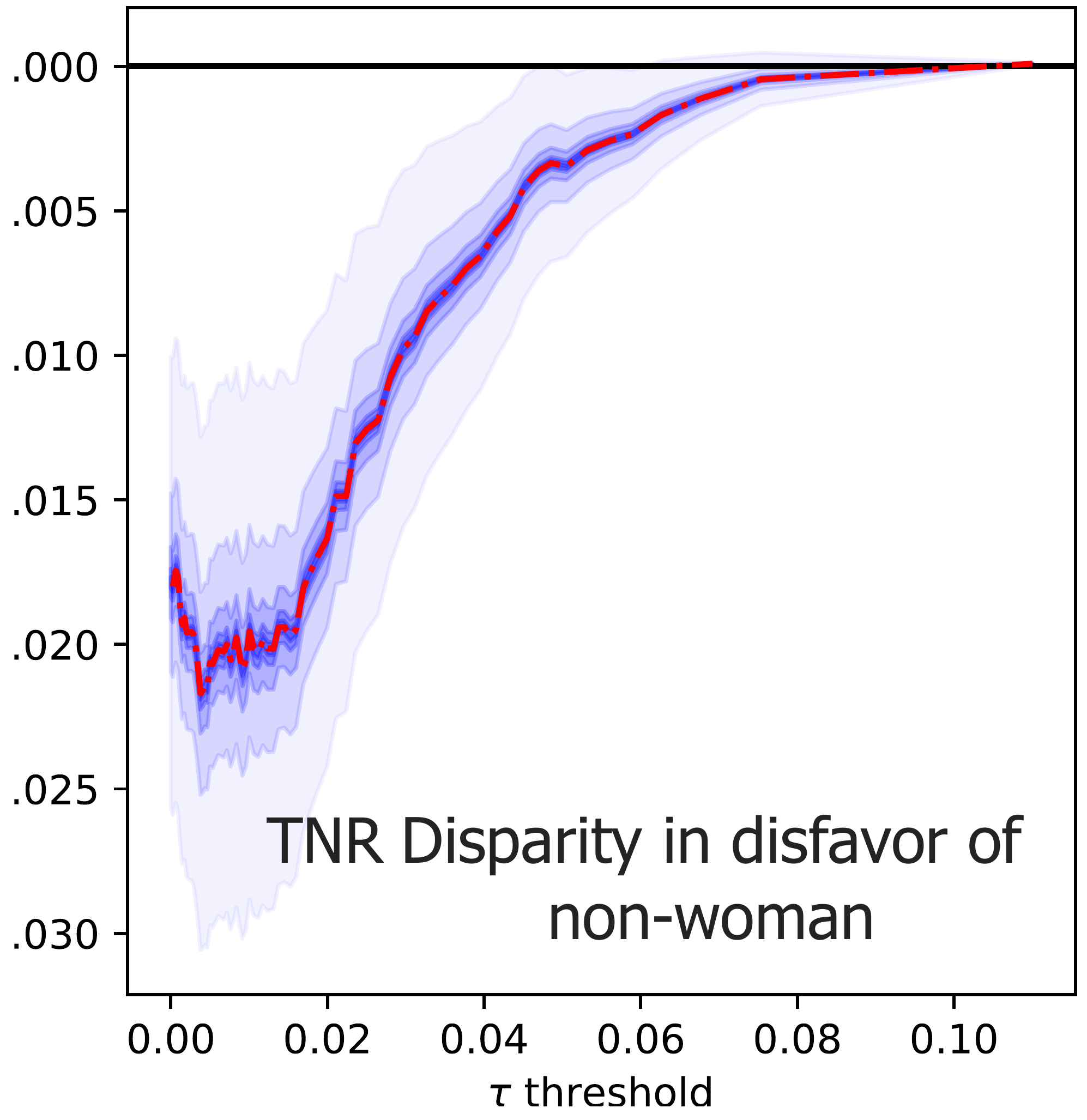}
    \includegraphics[width=0.4\linewidth]{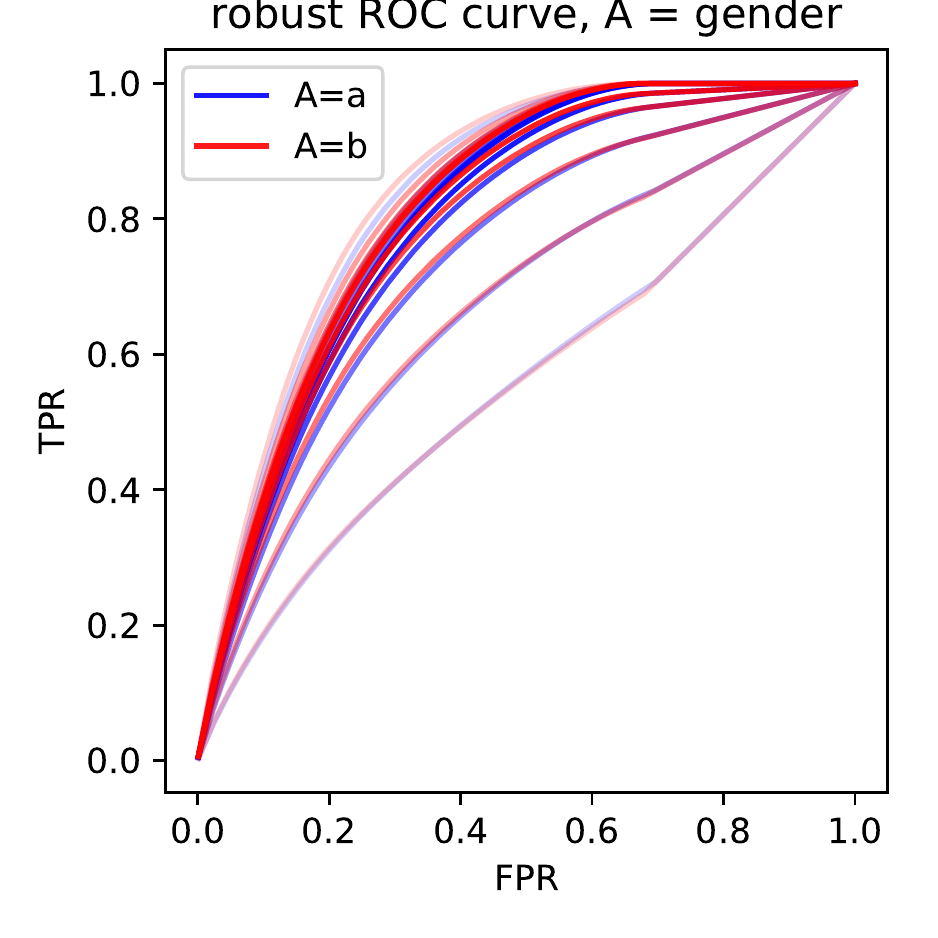}\hspace{0.1\linewidth}\includegraphics[width=0.4\linewidth]{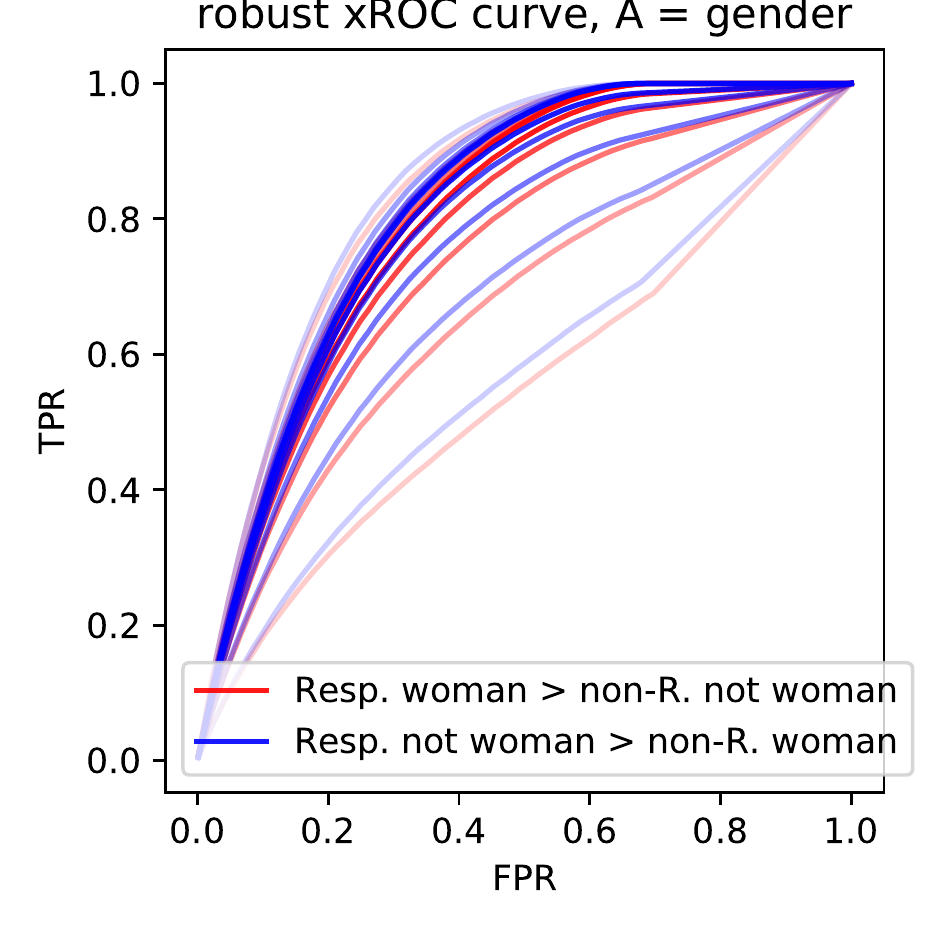}
		\caption{Diagnostics for gender protected attribute for \cref{sec-behaghel} (not-woman vs. woman)}
	\end{figure}
		\begin{figure}\centering
		\includegraphics[width=0.33\linewidth]{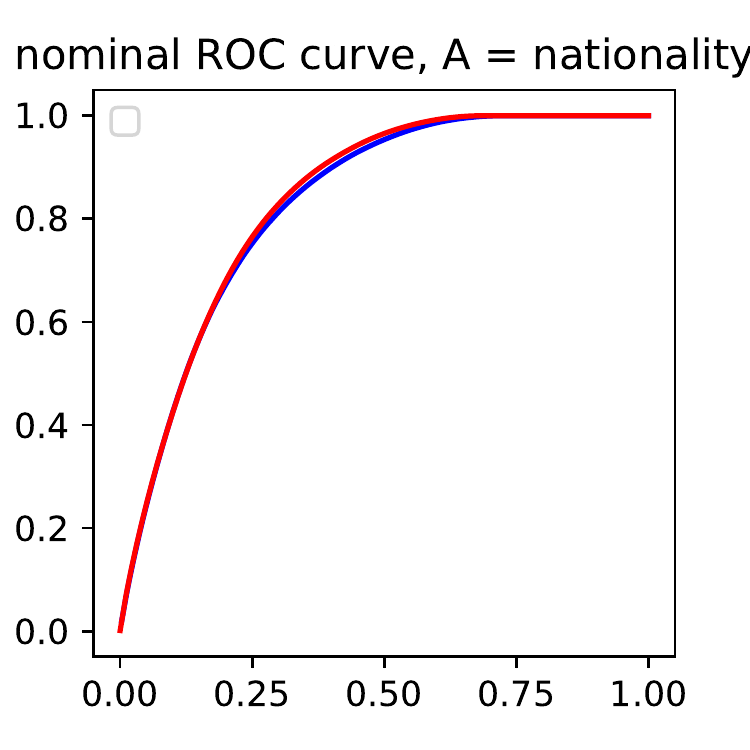}\includegraphics[width=0.33\linewidth]{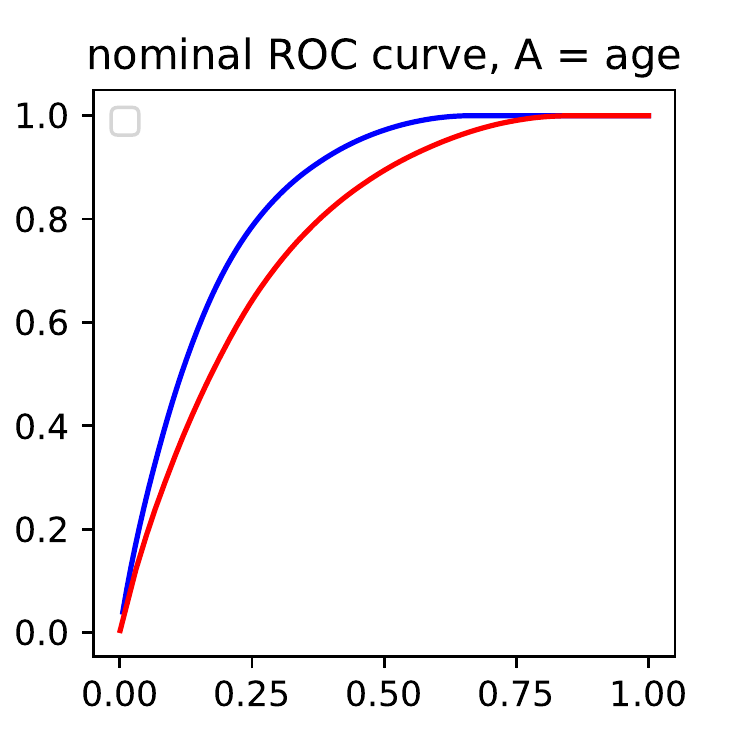}\includegraphics[width=0.33\linewidth]{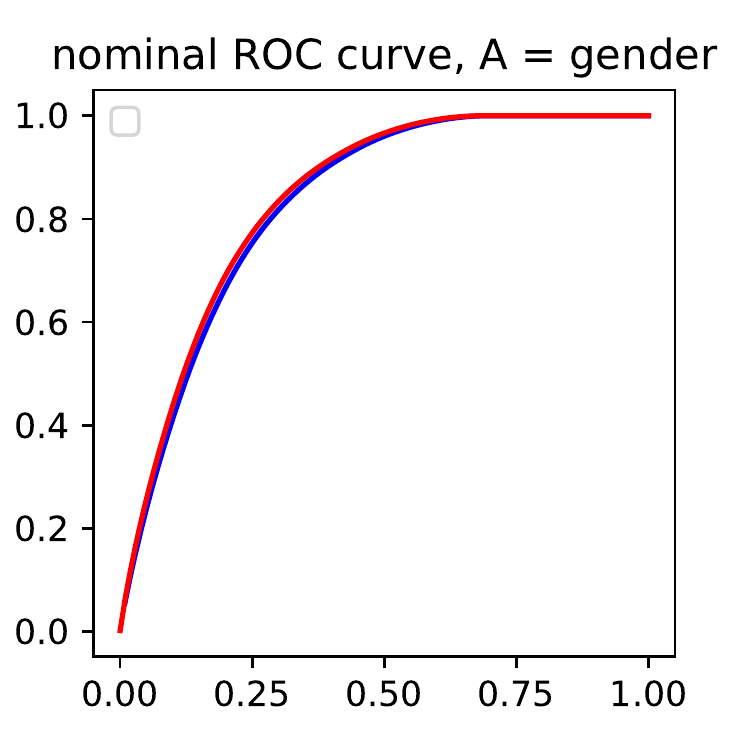}
		\caption{ROC curves under \cref{asm:monotone} for \cref{sec-behaghel} }
	\end{figure}
	
	\section{Substantive Discussion: Fairness vs. Justice}\label{apx-substantive}
We first caveat our use of ``disparate impact'': while our selection of protected attibutes parallels
	choices of protected attributes that appear elsewhere in the literature on fair machine learning, for the case of interventions,
	there may not be precedent from discrimination case law, nonetheless assessing fairness with respect to these social groups may be of concern. We view disparate impact in this domain as assessing fairness of outcome rates 
	under a personalization model. 
	
	\paragraph{Should true positive rates be adjusted for?}
	 Our presentation of an identification strategy of fairness %
	metrics for allocating interventions with unknown causal effects begs the question: should disparities in TPR and FNR be adjusted for in the interventional welfare setting? Is responder-accuracy parity a meaningful prescriptive notion of fairness? 
	
	One critique of outcome-conditional fair classification metrics recognizes the dependence of false positive rates on the underlying \textit{base rate}, $\pr(Y=1\mid A=a)$, \cite{cdg-18,c16}. The equivalent situation occurs when the within-group ATE varies by the protected attribute, e.g. $\E[\tau \mid A=a]$ differs. 

	Ultimately, external domain knowledge is required to %
	adjudicate
	whether group-wide disparities in ATE should be adjusted for, or to decide which normative notion of distributive justice or fairness is appropriate.
	For example, consider the case of job training. From an economic perspective, multiple mechanisms could explain heterogeneity in CATE by race. Active labor market programs (see \citep{labor-cb16}) may be less effective 
	for one group vs. another group due to the presence of labor-market discrimination. Alternatively, they could be less effective due to correlation of group status and efficacy that is mediated by occupation choice: one group may be more interested in labor markets where the primary benefits of job search counseling, in reducing search frictions, are not barriers to employment in the first place relative to other factors such as skills gaps. Intuitively, the former mechanism of ATE variation by group reflects a notion of ``disparity'' which remains problematic, while the latter may seem to reflect an unproblematic causal mechanism. 
	While mediation analysis and fairness defined in terms of path-specific effects could further decompose the treatment effect along these stated mechanisms, in policy settings, collecting all of the relevant information can be burdensome, and deciding on a causal graph can be difficult. 
	
	\textbf{Claims Across Outcomes}
	We first outline different frameworks for thinking about fairness/equity of algorithms and interventions. 
	Analogous to the proposals arising from metrics proposed in fairness in machine learning, one might view the decision-maker's concern to be of ensuring \textit{accuracy} parity, that the decisions meted out are overall beneficial to individual. 
	We view a theory of fairness that assesses disparities in outcome-conditional error rates in the context of a theory of normative claims arising from ``claims across outcomes''. \cite{adler} develops a ``claims across outcomes'' framework of fairness and social welfare, in the context of an overall welfarist theory of justice. 
	
	On the one hand, fair classification from the point of view of assessing or equalizing TPR or TNR disparities may be interpreted in a claims context as: for an individual with ``true outcome'' $Y$ and covariates $X$, an individual with the true label $Y=1$ as having a comparative claim for $\hat{Y}=1$, if the predictor $\hat{Y}$ is an allocation tool. We can map the setting of personalized interventions to the ``claims across outcomes'' setting: the potential outcomes framework posits for each individual the random variables of outcomes $Y(0), Y(1)$.  In the responder setting, the true label is responder status $Y(1) > Y(0)$. However, since these are \textit{jointly unobservable}, in situations where heterogeneous treatment effects are plausible, the best guess is an individual-level treatment effect conditional on covariates, $\mathbb{E}[Y(0)\mid X=x], \E[Y(1)\mid X=x]$. In this interventional setting, one can think of individuals having claims in favor of favorable outcomes, e.g. a claim in favor of $Y(1)$ if $Y(1) > Y(0)$. 
	
		For the case of interventions, classification decisions  $Z$ are allocative of real interventions, and we argue that implicitly, the consideration of social welfare (balancing efficiency and program costs) is an important factor in the original design of social programs or personalized interventions. This is in sharp contrast to the literature on fair classification which considers settings such as lending in finance, or risk prediction in the criminal justice system, where overriding concerns are primarily those of \textit{vendor} utility. 
	
	On the other side of the spectrum, we can recall axiomatically justified social welfare functions that apply to the case of \textit{deterministic} resource allocation, where outcomes are generally known. A decision-maker might also be concerned with equity considerations, adopting a min-max welfare criterion, appealing to Rawlsian justice frameworks. Another approach is simply assessing the population cardinal welfare of the allocation, e.g. the policy value $\E[Y(\pi(X))]$ or a social-welfare transformation thereof, $\E[g(Y(\pi(X)))]$. The literature on policy learning addresses welfare functionals that are linear functionals of potential outcomes, see \cite{kt15}. Cardinal welfare constraints such as those studied in \cite{heidari2018fairness} can be applied with an imputed CATE function. 
	\paragraph{Comparison to other work on fair classification and welfare.}
	 \cite{liu-drsh18} study the implications of classifier-based decisions, as well as proposals for statistical parity, on group welfare. Their work addresses selection rules that have known marginal impacts by group. \cite{hu2019fair} studies the welfare weights implied by classification parity metrics and shows that enforcing classification parity metrics are not Pareto-improving. Rather than studying the welfare implications of classification parity, we are concerned with assessing non-identifiable model errors in the causal-effect personalized intervention setting. %
	 Since in the personalized intervention setting, welfare is a primary objective for the Planner (e.g. social services, or social protection more broadly), modulo cost considerations, combining the distributional information from identification of classification errors with other social welfare objectives is of possible interest. 

We next aim to provide concrete examples of discussions regarding the distributional impacts of interventions, in order to provide additional context on different contexts wherein different notions of ``fairness'' from the fairness in machine learning literature map onto welfare or justice concerns, as stated in discussions on interventional outcomes. 
	
	\textbf{Lexicographic fairness or maximin (Rawlsian) fairness.}
	
	In a large multi-site graduation trial on testing an intensive, composite intervention targeted at the "ultra-poor", which comprised wraparound services including coaching and revenue-generating resources, still the poorest seemed to benefit least from the intervention in terms of sustained revenue \cite{banerjee2015multifaceted}. In this setting, concerns about maximin fairness (Rawlsian justice) might override considerations of efficiency insofar as one might be willing to invest resources to help the worst-off on humanitarian grounds. 
	
	\textbf{Universalism.}
	
	Criticisms of targeted policies in general note practical difficulties introduced by imposing and enforcing eligibility guidelines. \cite{m05}. Although discussion of resource constraints may be used to justify a targeting scheme, critics of targeting argue that the most efficient targeting is not as welfare-improving as simply advocating for greater resources \cite{eubanks2018automating}.

	\textbf{Additional distributional preferences on \( Y(Z)\) with respect to equitable or redistributive aims of the policy.}
	
	\cite{berger} consider profiling based on covariates as a means of allocating government services, in the example of allocating predicting unemployment duration to allocate reemployment services. They outline competing equity vs. efficiency concerns, in the case that unemployment duration is correlated with treatment efficacy (e.g efficacy of reemployment services), and conclude that `` tradeoffs between alternative social goals in designing profiling systems are likely to be empirically important... the form and extent of these tradeoffs may depend on empirical relationships between the impacts of the program being allocated and the equity-related characteristics of potential participants." While outcome-conditional true positive rates or true negative rates compare model performance across binary protected attributes, program designers may remain concerned regarding the distribution of benefits. \cite{carneiro2002removing} consider ``removing the veil of ignorance'' under the simplifying of constant treatment response to consider distributional (quantile) treatment effects, as a relaxation of the anonymity axiom of cardinal social welfare. Distributional preferences are relevant when program designers are concerned about model performance at finer-grained levels than discrete protected attribute.

\end{document}